\documentclass[letterpaper, 10 pt, conference]{ieeeconf}  

\IEEEoverridecommandlockouts                              

\usepackage[hidelinks]{hyperref}       
\usepackage{url}            
\usepackage{graphicx} 
\usepackage{amsmath} 
\usepackage{amssymb}  
\usepackage{paralist}
\usepackage{blindtext}
\usepackage{algorithm}
\usepackage{algpseudocode}
\usepackage{multirow}
\let\labelindent\relax 
\usepackage{enumitem}
\usepackage{tablefootnote}
\usepackage{booktabs}

\usepackage{array}
\newcolumntype{L}[1]{>{\raggedright\arraybackslash}p{#1}}
\newcolumntype{C}[1]{>{\centering\let\newline\\\arraybackslash\hspace{0pt}}p{#1}}
\newcolumntype{R}[1]{>{\raggedleft\arraybackslash}p{#1}}

\title{\LARGE \bf
A Multimodal Handover Failure Detection Dataset and Baselines
}

\author{Santosh Thoduka$^{1,3}$ and Nico Hochgeschwender$^{2}$ and  Juergen Gall$^{3,4}$ and Paul G. Pl\"{o}ger$^{1}$
\thanks{This work is supported by the European Union's Horizon 2020 research and innovation program under grant agreement No. 871252 (METRICS), the Deutsche Forschungsgemeinschaft (DFG, German Research Foundation) - GA 1927/4-2 (FOR 2535), the ERC Consolidator Grant FORHUE (101044724), and the Graduate Institute at Hochschule Bonn-Rhein-Sieg.}
\thanks{$^{1}$Hochschule Bonn-Rhein-Sieg, Germany
        {\tt\small \{santosh.thoduka, paul.ploeger\}@h-brs.de}}%
\thanks{$^{2}$University of Bremen, Germany
        {\tt\small nico.hochgeschwender @uni-bremen.de}}%
\thanks{$^{3}$University of Bonn, Germany
        {\tt\small gall@iai.uni-bonn.de}}%
\thanks{$^{4}$Lamarr Institute for Machine Learning and Artificial Intelligence, Germany}%
}

\begin{document}
\thispagestyle{empty}
\begin{minipage}[t]{\textwidth}
\textcopyright 2024 IEEE.  Personal use of this material is permitted.  Permission from IEEE must be obtained for all other uses, in any current or future media, including reprinting/republishing this material for advertising or promotional purposes, creating new collective works, for resale or redistribution to servers or lists, or reuse of any copyrighted component of this work in other works. The published version can be found at \url{https://doi.org/10.1109/ICRA57147.2024.10610143}
\end{minipage}
\newpage

\maketitle
\thispagestyle{empty}
\pagestyle{empty}

\begin{abstract}
An object handover between a robot and a human is a coordinated action which is prone to failure for reasons such as miscommunication, incorrect actions and unexpected object properties.
Existing works on handover failure detection and prevention focus on preventing failures due to object slip or external disturbances.
However, there is a lack of datasets and evaluation methods that consider \emph{unpreventable} failures caused by the human participant.
To address this deficit, we present the multimodal Handover Failure Detection dataset, which consists of failures induced by the human participant, such as ignoring the robot or not releasing the object.
We also present two baseline methods for handover failure detection:
\begin{inparaenum}[\itshape (i)\upshape]
    \item a video classification method using 3D CNNs and
    \item a temporal action segmentation approach which jointly classifies the human action, robot action and overall outcome of the action.
\end{inparaenum}
The results show that video is an important modality, but using force-torque data and gripper position help improve failure detection and action segmentation accuracy.
\end{abstract}

\section{Introduction}
An object handover between a human and a robot is a common action during tasks involving human-robot collaboration and physical human-robot interaction.
It requires both the giver and the receiver to communicate and coordinate their actions to ensure a safe and successful handover.
Failures in object handovers are likely, even for human-human handovers, for example, due to miscommunication, incorrect interpretation of intentions, object properties, or improper behaviour of either participant.
Detecting failures enables robots to react appropriately - either by attempting to recover from the failure or informing the person about the failure.
Certain types of failures, such as dropped objects, might cause harm, and thus detecting and handling them is important from a safety perspective.

Current approaches to error handling and failure detection in object handovers focus on the \emph{approach} and \emph{transfer} phase of the interaction, such as detecting slipping objects~\cite{parastegari2018failure}, increasing robustness to unwanted disturbances~\cite{davari2019identifying} and adapting to perception uncertainties~\cite{liu2022task}.
Other approaches include using tactile and force-sensing to reliably release the object when pulled~\cite{eguiluz2017reliable}, reactive handovers for unknown objects~\cite{yang2021reactive}, adapting to the human's motion during the handover~\cite{mavsar2022rovernet}, and estimating the properties of the container being received and appropriate grasp locations~\cite{pang2021towards}.
While these approaches are concerned with improving the reliability of handovers and preventing failures, in some cases, failures are not preventable, particularly when they are caused by actions of the human participants.
Therefore, failure \emph{detection} should be considered in conjunction with failure \emph{prevention} strategies.
\begin{figure}[tpb]
   \centering
   \includegraphics[width=0.85\linewidth]{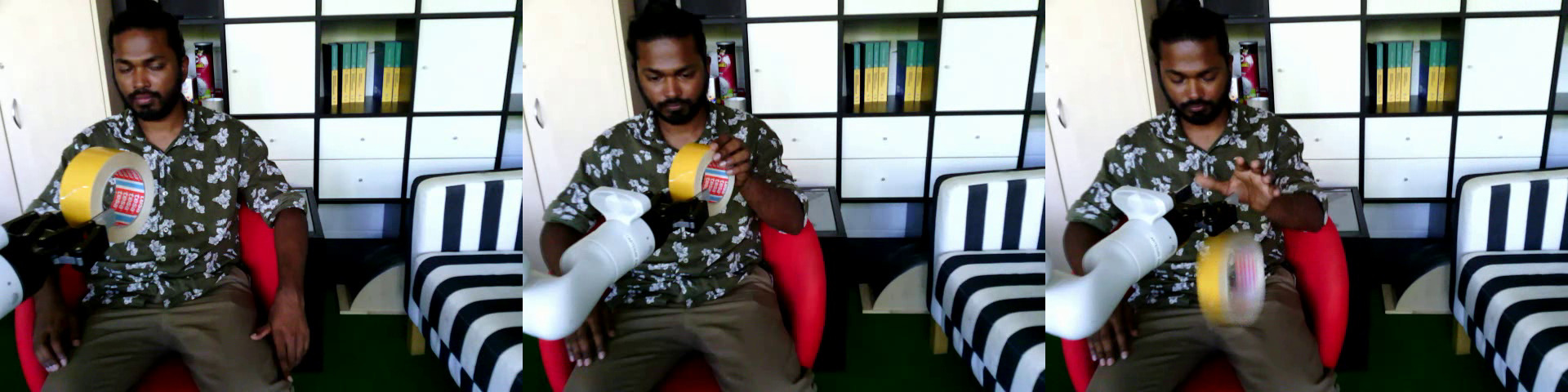}
   \includegraphics[width=0.85\linewidth]{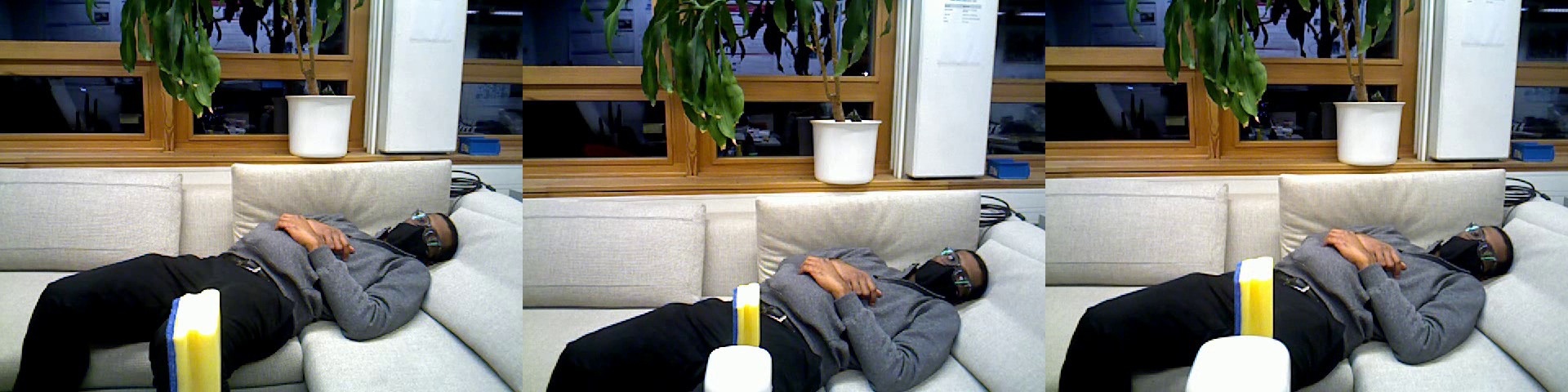}
    \caption{Examples of failed robot-to-human object handovers due to a dropped object (top) and a person ignoring the robot (bottom).}
   \label{fig:dropexample}
   \vspace{-4mm}
\end{figure}
Introducing failure conditions is an important aspect of benchmarking robots and evaluating their performance in realistic conditions.
Similar to software testing, subjecting a robot to abnormal conditions and injecting faults during a benchmark can be used to build arguments regarding its reliability and fault tolerance.
In the METRICS HEART-MET competition for healthcare robots\footnote{\url{https://metricsproject.eu/healthcare}}, benchmarks are developed to specifically include failure modes as independent variables in the benchmarking protocol~\cite{thoduka2021benchmarking}.
Failures are of particular relevance in healthcare contexts since the human partner might have cognitive or physical impairments, leading to failures such as dropped objects, as shown in Fig~\ref{fig:dropexample}.

Datasets that include failures in robotics are quite rare, partly because it is difficult to collect such datasets.
Failures typically have to be induced in order to balance the normal and failed instances since natural failures might occur rarely, and some types of failures might damage the robot or objects, or cause harm to people.
\begin{table*}[thpb]
  \caption{Objectives of the giver and receiver during R2H and H2R handovers and the associated failure modes}
  \label{tbl:failuremodes}
  \centering
    \begin{tabular}{l l  l  l}
    \toprule
     &  \textbf{Phase} & \textbf{Objective} & \textbf{Failure modes} \\
    \midrule
    \multirow{6}{*}{\rotatebox[origin=c]{90}{R2H}}     & \multirow{2}{*}{Approach} & the giver approaches the receiver with the object & \\
      &   &         the receiver approaches the giver & the human does not approach the robot\\
    \cmidrule(lr){2-4}
        & \multirow{2}{*}{Transfer} & the receiver grasps the object  &the human does not grasp the object \\
                     &       & the giver releases the object once it is safe to do so & the object is dropped \\
    \cmidrule(lr){2-4}
        & \multirow{2}{*}{Retract} &  the giver releases the object and moves away from the receiver &  \\
            &    & the receiver continues to grasp the object &    the object is dropped \\
    \midrule
     \multirow{6}{*}{\rotatebox[origin=c]{90}{H2R}}   & \multirow{2}{*}{Approach} & the giver approaches the receiver with the object & the human does not approach the robot \\
             &    & the receiver approaches the giver & the object is dropped \\
   \cmidrule(lr){2-4}
        & \multirow{2}{*}{Transfer} & the receiver grasps the object & \\
                     &        & the giver releases the object once it is safe to do so & the object is dropped  \\
    \cmidrule(lr){2-4}
       & \multirow{2}{*}{Retract} & the giver releases the object and moves away from the receiver & the human does not release the object \\
             &   & the receiver continues to grasp the object & \\
    \bottomrule
  \end{tabular}
  \vspace{-4mm}
\end{table*}

In this paper, we introduce the Handover Failure Detection (HFD) dataset in order to address the requirement for failure detection benchmarks for object handovers.
The types of handover failures in the dataset formed part of the METRICS HEART-MET competition at ICRA 2023 as possible behaviours of persons for whom a robot had to fetch an object.
The dataset has also been used to create a handover failure detection benchmark hosted on Codabench~\cite{codabench}. 
We focus on failures that are caused by the human participant in both robot-to-human (R2H) and human-to-robot (H2R) handovers.
The dataset includes multimodal data such as video, robot joint state, and readings from a force-torque sensor.
Multimodal data is important for handovers since both visual cues and physical signals (e.g. torque) influence the coordination~\cite{grigore2013joint}.
We also present two baseline approaches for detecting the failures: the first uses video classification with 3D CNNs and the second uses temporal action segmentation to jointly classify the actions of the human participant, actions of the robot, and the overall outcome of the trial, both of which show the importance of learning from multimodal data.

The contributions of this paper are:
\begin{inparaenum}
    \item the multimodal HFD dataset consisting of human-induced failures during R2H and H2R handovers and
    \item baseline methods for detecting handover failures and recognizing the type of failure using video classification and action segmentation methods\footnote{The dataset, code and Codabench benchmark are available at: \url{https://sthoduka.github.io/handover_failure_detection}}.
\end{inparaenum}


\section{Related Work}
The survey on object handovers by Ortenzi et al.~\cite{ortenzi2021object} identifies works which consider error handling during handovers; most approaches that consider failures in handovers typically focus on the physical handover phase, which is concerned with a failure to coordinate between the receiver and the giver.
For example, both Eguiluz et al.~\cite{eguiluz2017reliable} and Davari et al.~\cite{davari2019identifying} try to differentiate between an object being pulled (as part of the nominal R2H task) versus other disturbances, so that the robot does not release the object unexpectedly.
Similarly, Parastegari et al.~\cite{parastegari2016fail,parastegari2018failure} detect slipping objects and regrasp them to prevent failures caused by objects falling.
Rosenberger et al.~\cite{rosenberger2020object} prevent failures by aborting the handover if certain error conditions are met (such as the human moving too close to the grasp position, or if a collision is detected).
Overall, the survey reports that only a minority of approaches incorporate error handling in their work, with the primary focus being to detect disturbances and preventing falling objects.

Liu et al.~\cite{liu2022task} describe approaches to increase safety by adapting the behaviour of the robot depending on the uncertainty of its perception systems.
For example, to prevent collisions in darker environments, the robot maintains a larger distance to the human since the uncertainty of their skeleton detection system increases.
Han and Yanco~\cite{han2020reasons} consider unrecoverable handover failures in their work, but investigate the expectations that a person has when a failure occurs.
They find that people expect the robot to provide explanations, and not just non-verbal cues, for why a failure has occurred.
In other work~\cite{meng2022fast, mamaev2021grasp}, specific failures, such as occlusions or no detections of the human's hands, are considered to either disregard an experimental trial, or to abort the handover.
Iori et al.~\cite{iori2023dmp} consider two scenarios which deviate from a nominal handover, in which
\begin{inparaenum}[\itshape (i)\upshape]
    \item the person pauses while approaching the robot before continuing, and
    \item the person's arm is pushed by another person during the approach.
\end{inparaenum}
The paper presents a reactive trajectory generation method which is able to adapt to the perturbations.
Grigore et al.~\cite{grigore2013joint} aim to increase safety in R2H by considering human engagement signals (e.g.\ gaze) in addition to physical signals (e.g.\ current and torque) to decide if and when to release the object, and Mohandes et al.~\cite{mohandes2022robot} use vision and joint torques to detect the person's intentions and actions. 

Prior research related to handover failures are concerned with preventing failures, or detecting conditions which might lead to a failure, and typically only deal with failures in the pre-handover and the physical handover phases.
The post-handover phase should also be considered as it might take some time for the receiver to stabilize the object in their hand and get accustomed to its physical properties such as temperature or weight.
In our work, we are concerned with detecting failures after they have occurred, with a specific emphasis on human-induced failures.
We also consider failures in the post-handover phase for both H2R and R2H.

\section{The Handover Failure Detection Dataset}
\begin{figure*}[thpb]
   \centering
   \includegraphics[width=0.9\linewidth]{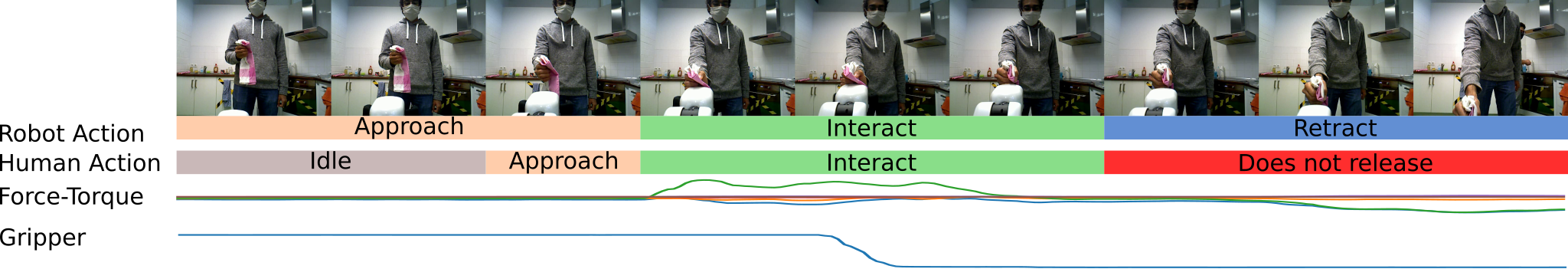}
    \caption{The dataset consists of video, force-torque readings and robot joint states, and contains annotations for the robot and human actions.}
   \label{fig:human_robot_activity}
    \vspace{-4mm}
\end{figure*}

In this section, we introduce the HFD dataset, which consists of successful and failed R2H and H2R object handovers.
The primary motivation for creating the dataset is to improve the monitoring capabilities of robots, thus enabling them to react to failure conditions appropriately.

\subsection{Development}
The failures in the dataset are induced by the participants based on prior instructions.
We select the failures by following the procedure defined in~\cite{thoduka2021benchmarking} for benchmarking robots by inducing failure scenarios.
The procedure requires defining the objective of the task and identifying failure modes, which may be associated with negations of the objective.
The objective of a handover task is to ``\emph{safely transfer an object from the giver to the receiver}"; this is further decomposed into the objectives of the three phases of a handover, \emph{approach}, \emph{transfer}, \emph{retract}, as shown in Table~\ref{tbl:failuremodes}.
The failure modes in Table~\ref{tbl:failuremodes} are identified by considering the objective of the human participant and specifying conditions under which the objective is not satisfied.
Other independent variables, which are varied for each trial, include the robot platform, person, person's pose, object, and location.

We use real robots and human participants instead of simulation to provide a more realistic benchmark.
In the future, simulated data can be used as additional training data.
The HandoverSim framework~\cite{chao2022handoversim} is a promising approach in this direction, where the simulation of the human's hand is derived from a motion capture dataset.
If this is extended to simulate an entire human realistically, it would be possible to create large-scale training data in simulation where failures can be induced in a controlled manner.
Nevertheless, it is important to evaluate on real robots, which is the purpose of the HFD dataset. 

\subsection{Participant Recruitment}
We recruited a total of 17 participants, all of whom were students familiar with robots.
Participation was voluntary with no monetary or other compensation.
Informed consent for the use of their data was obtained from the participants before starting the data collection.

\subsection{Procedure}
For a given trial, the procedure is as follows:
\begin{inparaenum}[\itshape (i)\upshape]
    \item the participant is instructed before the start of the trial about their desired behaviour, which may or may not induce a failure;
    \item data recording is started and the robot executes the handover action;
    \item the robot releases or grasps the object if a threshold is reached on the force-torque sensor;
    \item the robot retracts its arm and data recording is stopped.
\end{inparaenum}

\subsection{Dataset Characteristics}
The dataset consists of 589 trials performed with two robots, 17 participants and 22 object classes.
The class distribution and distribution between handover type and robot platform is summarized in Table~\ref{tbl:datasetstats}.
The objects used include household objects such as a book, towel, sponge, bowl, bottle, etc.; a full list can be found in the datasheet which is included with the dataset.

\begin{table}[t]
  \caption{Dataset statistics}
  \label{tbl:datasetstats}
  \centering
    \begin{tabular}{@{}p{0.085\textwidth}R{0.012\textwidth}R{0.012\textwidth}R{0.012\textwidth}R{0.012\textwidth}R{0.012\textwidth}R{0.012\textwidth}R{0.012\textwidth}R{0.012\textwidth}R{0.012\textwidth}R{0.012\textwidth}@{}}
    \toprule
        & \multicolumn{5}{c}{R2H} & \multicolumn{5}{c}{H2R}\\
    \cmidrule(lr){2-6}\cmidrule(lr){7-11}
        & \multicolumn{1}{c}{\rotatebox{90}{success}} & \multicolumn{1}{c}{\rotatebox{90}{no approach}} & \multicolumn{1}{c}{\rotatebox{90}{no grasp}} & \multicolumn{1}{c}{\rotatebox{90}{drop}} & \multicolumn{1}{c}{\rotatebox{90}{\textbf{total}}} & \multicolumn{1}{c}{\rotatebox{90}{success}} & \multicolumn{1}{c}{\rotatebox{90}{no approach}} & \multicolumn{1}{c}{\rotatebox{90}{no release}} & \multicolumn{1}{c}{\rotatebox{90}{drop}} & \multicolumn{1}{c}{\rotatebox{90}{\textbf{total}}} \\
    \midrule
        Toyota HSR & 68 & 50 & 49 & 58 & 225 & 51 & 46 & 57 & 66 & 220 \\
    
        Kinova Gen3 & 18 & 17 & 17 & 20 & 72 & 19 & 18 & 17 & 18 & 72\\
    \midrule
        \textbf{Total} &  \textbf{86} & \textbf{67} & \textbf{68} & \textbf{76} & \textbf{297} & \textbf{70} & \textbf{64} & \textbf{74} & \textbf{84} & \textbf{292} \\
    \bottomrule
  \end{tabular}
  \vspace{-4mm}
\end{table}

The data for each trial includes RGB video from a robot-mounted camera, joint states of the robot arm (position, velocity and effort for each joint, including the gripper) and force-torque (F-T) sensor data\footnote{A wrist-mounted force-torque sensor on the Toyota HSR, and simulated force-torque readings at the wrist derived from the individual torques of the joints of the Kinova arm}.
In addition to the raw sensor data, resampled joint states and F-T data which are aligned to the timestamps of the video frames are included.

The training, validation and test sets are split such that a participant is present in only one of the sets, resulting in 337 trials for training, 101 for validation, and 151 for testing.
In the approaches presented in the next section, all results are reported on the test set, with the validation set being used for hyperparameter selection and for early stopping.
    \begin{figure*}[tpb]
       \centering
       \includegraphics[width=0.95\linewidth]{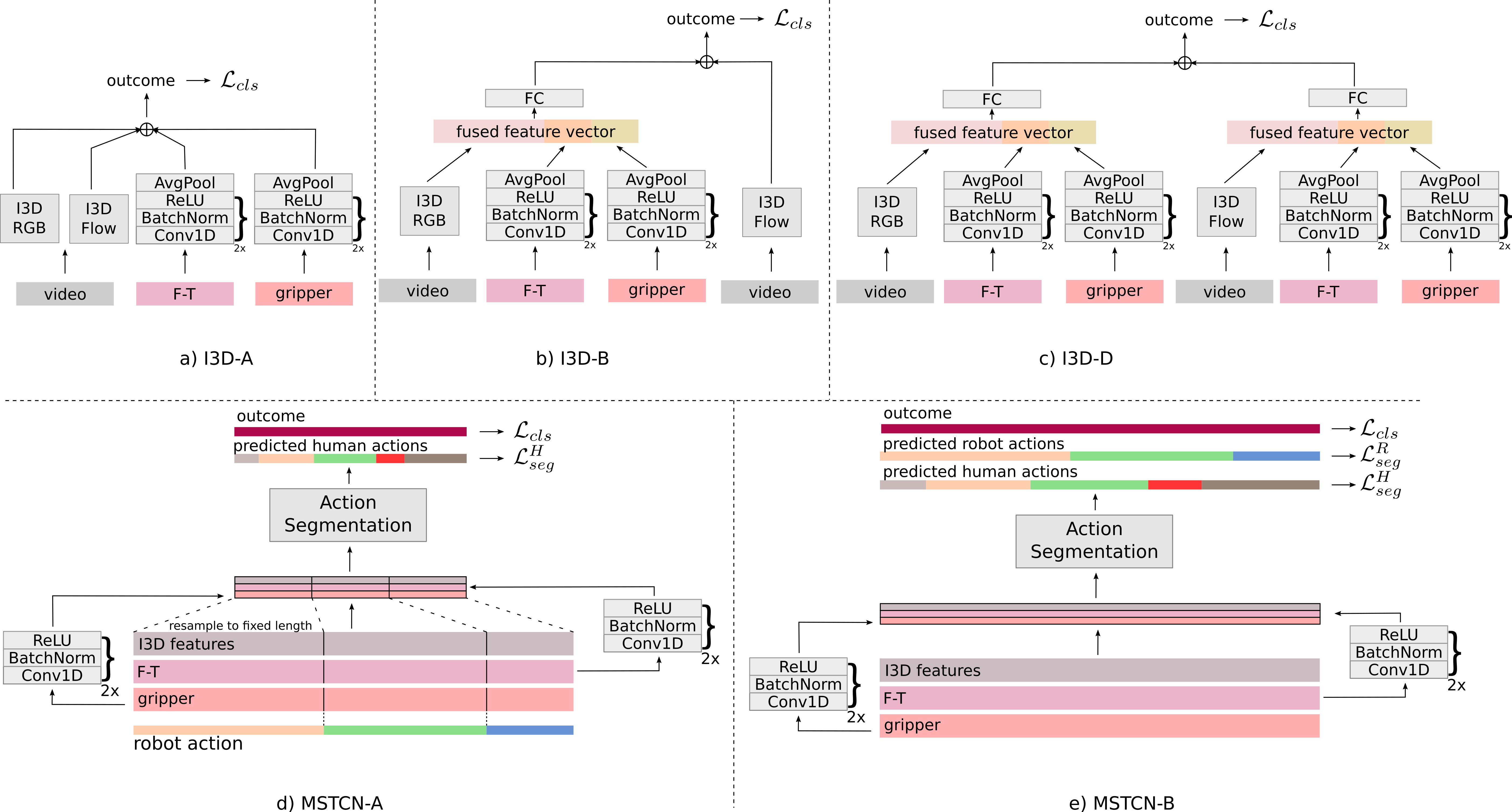}
        \caption{a) \textbf{I3D-A:} I3D network with late fusion of modalities. b) \textbf{I3D-B:} I3D network with intermediate fusion of F-T and gripper features with RGB stream. \textbf{I3D-C}, which is not shown, is very similar to I3D-B, but the intermediate fusion of F-T and gripper features is with the Flow instead of the RGB stream.
        c) \textbf{I3D-D:} I3D network with intermediate fusion of F-T and gripper features with RGB and Flow streams. d) \textbf{MSTCN-A:} MSTCN network that takes I3D, F-T and gripper features as input and predicts the human actions for each frame and the outcome. The input is resampled to a fixed length based on the robot actions. e) \textbf{MSTCN-B:} MSTCN network with original-length inputs and an additional prediction head for the robot actions.}
       \label{fig:networks}
       \vspace{-4mm}
    \end{figure*}
\subsection{Annotations}
The robot's actions (idle, approach, transfer, retract, post-idle) are extracted based on the robot's joint states\footnote{The robot actions always follow the same temporal order, therefore only the boundaries between the actions are determined based on whether the arm is moving or not. As such, the actions of the robot are assumed to be known and can also be retrieved from an action executor, such as a state machine.}.
The person's actions are manually annotated, including the handover phases and non-nominal actions (idle, approach, transfer, retract, post-idle, not released, dropped). 
The start and end of each action is annotated, such that each frame of the video is associated with a robot action and a human action (see example in Fig.~\ref{fig:human_robot_activity}).
The transfer phase was annotated by observing the F-T data and video to determine the start and end of the physical interaction of the person with the object and robot.
The start of the approach phase and the end of the retract phase were annotated by observing the motion of the person's arm in the video\footnote{Although all annotations were performed by a single annotator, there might still be some ambiguity in the exact boundaries of the person's actions.}.
The outcome of the trial is one of the classes listed in Table~\ref{tbl:datasetstats}.
The robot (Toyota Human Support Robot (HSR) or Kinova Gen3) and the task (R2H or H2R) are included as metadata for each trial.

\section{Baseline Approaches}
The input data consists of a sequence of \(T\) video frames \(x_{1:T} = (x_1, ..., x_T)\), force-torque measurements \(ft_{1:T} = (ft_1, ... , ft_T)\)\footnote{\(ft\) is a vector consisting of the force \((f_x, f_y, f_z)\) and torque \((\tau_x, \tau_y, \tau_z)\) in three directions expressed in Newton and Newton-meter, respectively.} and the state of the gripper \(g_{1:T} = (g_1, ... g_T)\), where \(g_t \in\) \{-0.5, 0.0, 0.5\} refers to \{open, partially closed, and closed\}, respectively.
Additional annotations are the current human action \(h_{1:T} = (h_1,...,h_T)\) being performed, where \(h_t \in\) \{idle, approach, interact, retract, post-idle, not released, dropped\} and which is only available during training, and the current robot action \(r_{1:T} = (r_1, ..., r_T)\) being performed, where \(r_t \in\) \{approach, interact, retract\} and which is available during training and inference time.
The task is to classify the outcome of the handover action \(o \in\) \{success, no approach, no grasp, drop, no release\}, where \emph{no grasp} only applies to R2H and \emph{no release} only applies to H2R.
We present two baseline methods for detecting the outcome of the handover action.
    \subsection{Video classification}
    For the first baseline, we propose multi-modal variants of the pre-trained Inflated 3D ConvNet (I3D)~\cite{carreira2017quo} network.
    We investigate four variants, which are illustrated in Fig.~\ref{fig:networks}:
    \begin{enumerate}[label=(\alph*)]
        \item \textbf{I3D-A}: networks to classify the outcome from video (with separate RGB and optical flow streams), F-T and gripper state are trained separately, and their outputs are combined with late fusion.
        \item \textbf{I3D-B}: intermediate fusion and a fully-connected (FC) layer are used to combine the features of the RGB stream, F-T and gripper, whose output is combined with the output of the optical flow stream with late fusion.
        \item \textbf{I3D-C}: intermediate fusion and a FC layer are used to combine the features of the optical flow stream, F-T and gripper, whose output is combined with the output of the RGB stream with late fusion.
        \item \textbf{I3D-D}: intermediate fusion is used to combine features of the F-T and gripper streams with both RGB and optical flow features, before applying late fusion to combine the two outputs. 
    \end{enumerate}
    For the video stream, the input consists of 64 frames that are sampled at equal intervals from the entire video of a trial.
    F-T measurements are normalized based on the standard deviation and mean of measurements in the training set.
    
    \subsection{Human action segmentation}
     \begin{figure}[tpb]
       \centering
       \includegraphics[width=0.88\linewidth]{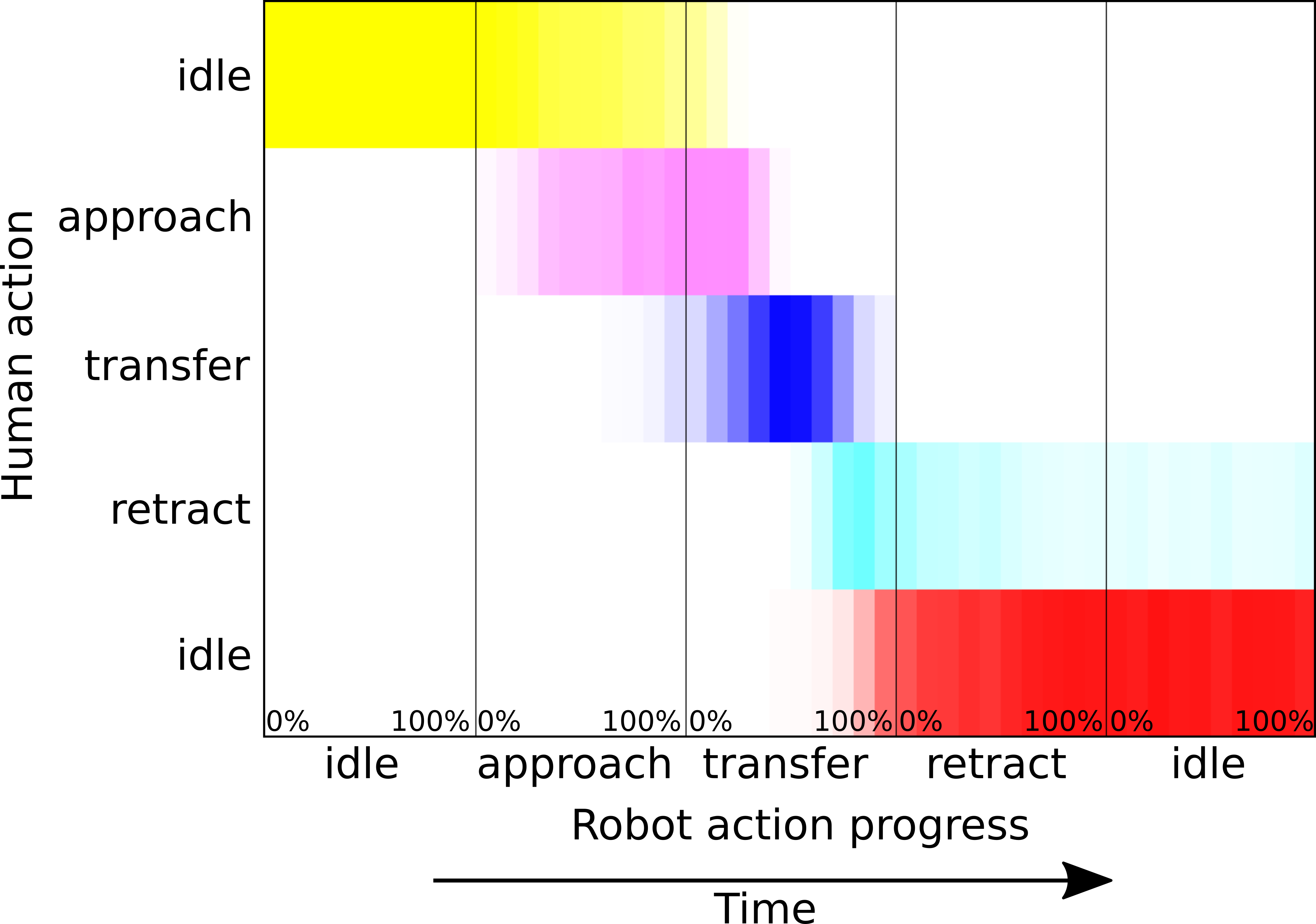}
        \caption{Frequency of each human action corresponding to the robot's actions in the training set for nominal R2H handovers. The progress of each robot action from 0\% - 100\% is discretized into 10 bins}
       \label{fig:activity_corr}
       \vspace{-6mm}
    \end{figure}
    
    The failures in the dataset are caused either by the person not performing the expected action (such as not reaching out) or performing an unexpected action (such as dropping the object).
    Therefore, we consider the auxiliary task of classifying the actions of the human as a means of detecting unperformed and unexpected actions.
    Recent approaches for temporal action segmentation in untrimmed videos make use of temporal convolutions~\cite{lea2017temporal, farha2019ms, li2023ms} or transformer-based architectures~\cite{yi2021asformer}.
    We select the multi-stage temporal convolutional network (MS-TCN)~\cite{farha2019ms} due to its good performance on action segmentation datasets and since transformer-based methods require larger datasets.
    Although improved versions of MS-TCN exist~\cite{ding2023temporal}, we use the basic architecture here for the baseline.
    
    MS-TCN uses dilated temporal convolutions on a sequence of clip-level features to perform action segmentation.
    Each stage of the network outputs a prediction, which is refined by the next stage.
    We use a 2-stage MS-TCN network for segmenting the actions of the human and extend it to the multi-modal setting.
    RGB and optical flow features are extracted for each frame from pre-trained I3D networks using a 64-frame window from the past, resulting in a feature vector of length 2048 for each frame.
    F-T and gripper state are passed through 1D convolutional layers before being concatenated to the I3D features.
    Therefore, the input to the segmentation network is a sequence of feature vectors \(f_{1:T} = (f_1,...f_T)\), representing the video, F-T and gripper state.
    The target for the network is the sequence of human actions \(h_{1:T}\) and the overall outcome \(o\).
    The predicted overall outcome is output as a sequence \(\hat{o}_{1:T}\), although only \(\hat{o}_T\) is considered at test time. 

    Fig.~\ref{fig:activity_corr} shows histograms for each human action corresponding to the robot action being performed in nominal R2H handovers (from the training set only).
    It can be seen, for example, that the person typically approaches the robot when the robot is approaching the person and during the initial stages of \emph{transfer}.
    In order to exploit this correlation, we consider two ways in which the robot action can be used to improve the classification of the human action (see Fig.~\ref{fig:networks}):
    \begin{enumerate}[label=(\alph*)]
    \item \textbf{MSTCN-A}: the input features are resampled such that the features corresponding to each robot action are represented by a fixed length sequence.
    \item \textbf{MSTCN-B}: the original sequence length is used, but the network predicts the robot action in addition to predicting the human action and overall outcome.
    \end{enumerate}

    \subsubsection{Loss functions}
    The cross entropy loss is used for the classification loss of the outcome, \(\mathcal{L}_{cls} = -\log(\hat{o}_c)\), where \(\hat{o}_c\) is the predicted softmax score for the ground-truth class \(c\).
    For the segmentation network, the classification loss for the outcome is summed over all frames \(t\) and all stages \(s\) of the multi-stage network, so that \(\mathcal{L}_{cls} = -\sum_s \frac{1}{T} \sum_t \log(\hat{o}^{s,t}_c)\).
    
    The segmentation loss (Eq.~\ref{eq:seg}) consists of the classification loss of the robot or human action (Eq.~\ref{eq:segcls}), and a smoothing loss (Eq.~\ref{eq:smooth}) which reduces over-segmentation~\cite{farha2019ms}.
    The smoothing loss is weighted by \(\lambda = 0.15\) and we use \(\tau = 4\) as in~\cite{farha2019ms}.
    The human action and robot action segmentation loss terms, \(\mathcal{L}_{seg}^H\) and \(\mathcal{L}_{seg}^R\), are also summed over all stages \(s\) of the multi-stage network.
    \(\hat{y}^{s,t}_{c}\) is the probability of class \(c\) at time \(t\) and stage \(s\), where \(c\) corresponds to the ground-truth human or robot action, respectively.
    \begin{align}
    \mathcal{L}_{segcls} &= -\sum_s \frac{1}{T} \sum_t{\log(\hat{y}^{s,t}_{c})} \label{eq:segcls}\\
    \mathcal{L}_{smooth} &= \sum_s \frac{1}{TC} \sum_{t,c} \min(|\log(\hat{y}^{s,t-1}_{c}) - \log(\hat{y}^{s,t}_{c})|, \tau)^2 \label{eq:smooth}\\
    \mathcal{L}_{seg} &= \mathcal{L}_{segcls} + \lambda \mathcal{L}_{smooth}  \label{eq:seg}
    \end{align}
    
    \subsubsection{Metrics}
    All networks are trained five times, and we report the mean and standard deviation of the classification accuracy for the outcome and the segmentation metrics, which are frame-wise accuracy and segmental F1-score~\cite{lea2017temporal} at overlapping thresholds of 10\%, 25\% and 50\%.

\section{Results}
\label{sec:exp}
The main results are shown in Table~\ref{tbl:experiments}, in which the networks are trained using different combinations of input modalities and loss functions.
For I3D-A with individual modalities (1-3), the video modality performs better than F-T or gripper alone.
In terms of fusion strategies (4-7), I3D-D, in which F-T and gripper features are combined using intermediate fusion with both RGB and flow features, performs best. It also outperforms the uni-modal variants. It needs to be noted that for a single modality, all I3D variants are the same.    
When we analyze the impact of each modality (8-12), we observe that the combination of video and F-T performs best.   
We next investigate multitask learning with MSTCN by considering only video as input (13-17). We observe a clear improvement when \(\mathbf{\mathcal{L}_{seg}^H}\) is added as a second task. Overall, MSTCN-B with all three tasks performs best. It also performs better than the video-only I3D-A model (1).
In the case of multimodal inputs (18-24), both MSTCN variants show the best performance with all three modalities, with a marginal drop in performance when using the robot action segmentation loss term for MSTCN-B (24). The latter is not surprising since the additional modalities provide information about the robot action such that predicting the robot action is not a challenging task anymore.   
We also note that the video + F-T combination (18, 21) performs better compared to video + gripper (19, 22), as with I3D-D (9, 10).

\begin{table}[thpb]
  \caption{Impact of modalities, loss terms and network}
  \label{tbl:experiments}
  \centering
   \addtolength{\tabcolsep}{-0.2em}
    \begin{tabular}{l | l | c  c  c | c  c  c | c }
    \hline
                    &   & \multicolumn{3}{c|}{\textbf{Modality}} & \multicolumn{3}{c|}{\textbf{Loss function}} & \\
    \hline
     \textbf{ID}  & \textbf{Model} & \textbf{V} & \textbf{F-T} & \textbf{G} & \(\mathbf{\mathcal{L}_{cls}}\) & \(\mathbf{\mathcal{L}_{seg}^H}\) & \(\mathbf{\mathcal{L}_{seg}^R}\) & \textbf{Accuracy}\\
    \hline
      1&  I3D-A & \checkmark & & & \checkmark & & & \textbf{64.8  \(\pm\) 3.3} \\
      2&  I3D-A &  &  \checkmark &  & \checkmark & & & 29.4  \(\pm\) 6.1 \\
      3&  I3D-A &  &  &  \checkmark & \checkmark & & & 38.2  \(\pm\) 7.5 \\
    \hline
      4&  I3D-A & \checkmark & \checkmark & \checkmark & \checkmark & & & 65.4  \(\pm\) 2.2 \\
      5&  I3D-B & \checkmark & \checkmark & \checkmark & \checkmark & & & 66.5  \(\pm\) 3.0 \\
      6&  I3D-C & \checkmark & \checkmark & \checkmark & \checkmark & & & 64.1  \(\pm\) 1.4 \\
      7&  I3D-D & \checkmark & \checkmark & \checkmark &  \checkmark & &  &   \textbf{67.9 \(\pm\) 3.3} \\
    \hline
      8&  I3D-D & \checkmark & & & \checkmark & & & 64.8  \(\pm\) 3.3 \\
      9&  I3D-D & \checkmark & \checkmark &  &  \checkmark & &     & \textbf{69.1 \(\pm\) 1.5} \\
      10&  I3D-D & \checkmark &  & \checkmark &  \checkmark & &     &   67.3 \(\pm\) 2.6\\
      12&  I3D-D & \checkmark & \checkmark & \checkmark &  \checkmark & &  &   67.9 \(\pm\) 3.3 \\
    \hline
      13&  MSTCN-A & \checkmark & & &  \checkmark &  &            &  57.9 \(\pm\) 1.8 \\
      14&  MSTCN-A & \checkmark & & &  \checkmark & \checkmark &            &  63.8 \(\pm\) 4.8 \\
      15&  MSTCN-B & \checkmark & & &  \checkmark &                        & & 56.6 \(\pm\) 2.2 \\
      16&  MSTCN-B & \checkmark & & &  \checkmark & \checkmark &            &  62.7 \(\pm\) 3.2 \\
      17&  MSTCN-B & \checkmark & & &  \checkmark & \checkmark & \checkmark &  \textbf{67.8 \(\pm\) 3.7} \\
    \hline
     18& MSTCN-A & \checkmark & \checkmark &  &  \checkmark & \checkmark  & &  67.3 \(\pm\) 2.3\\
     19&  MSTCN-A & \checkmark &  & \checkmark &  \checkmark & \checkmark  & &  65.6 \(\pm\) 2.0\\
     20&  MSTCN-A & \checkmark & \checkmark & \checkmark &  \checkmark & \checkmark  & &  \textbf{71.4 \(\pm\) 1.8}\\
     21&  MSTCN-B & \checkmark & \checkmark &  & \checkmark & \checkmark &  & 69.0 \(\pm\) 2.7 \\
     22&  MSTCN-B & \checkmark & &  \checkmark & \checkmark & \checkmark &  & 66.4 \(\pm\) 2.8 \\
     23&  MSTCN-B & \checkmark & \checkmark & \checkmark &  \checkmark & \checkmark &  &  \textbf{71.1 \(\pm\) 2.6}\\
     24&  MSTCN-B & \checkmark & \checkmark & \checkmark &  \checkmark & \checkmark & \checkmark &  70.7 \(\pm\) 1.8 \\
    \hline
    \multicolumn{8}{l}{\footnotesize{V: Video/I3D features, F-T: force-torque, G: gripper}}

  \end{tabular}
  \vspace{-4mm}
\end{table}

\begin{table}[thpb]
  \caption{Segmentation Metrics}
  \label{tbl:segmetrics}
  \centering
   \addtolength{\tabcolsep}{-0.2em}
    \begin{tabular}{l | c | c | c  | c }
    \hline
        \textbf{Model} &   \multicolumn{3}{c|}{\textbf{F1@\{10,25,50\}}} & \textbf{Frame-wise acc}\\
    \hline
       \multicolumn{5}{l}{\textbf{Baseline using correlation between robot and human action}} \\
       \hline
       Correlation & 58.1  & 48.6 & 30.8 & 44.5 \\
       \hline
       \multicolumn{5}{l}{\textbf{I3D features only}} \\
       \hline
       MSTCN-A &   74.7 \(\pm\) 1.4 & 68.0 \(\pm\) 1.3 & 51.1 \(\pm\) 1.9 & 70.5 \(\pm\) 2.6\\
       MSTCN-B &   67.9  \(\pm\) 2.9 & 62.7 \(\pm\)  3.0 & 48.2 \(\pm\) 2.2 & 69.6 \(\pm\) 2.3\\
       \hline
       \multicolumn{5}{l}{\textbf{I3D features, F-T and gripper}} \\
       \hline
       MSTCN-A &   80.3  \(\pm\) 0.5 & 74.1  \(\pm\) 0.7 & 57.1  \(\pm\) 2.3 & 75.9  \(\pm\) 1.1\\
       MSTCN-B &   70.8  \(\pm\) 2.0 & 65.2  \(\pm\) 2.1 & 49.7  \(\pm\) 2.6 & 73.2 \(\pm\) 0.8\\
       \hline
       \multicolumn{5}{l}{\textbf{I3D features, F-T and gripper (only \(\mathbf{\mathcal{L}_{seg}^H}\))}} \\
       \hline
       MSTCN-A &  74.7   \(\pm\) 2.2 & 70.7 \(\pm\) 1.9 &  56.7 \(\pm\) 1.5 & 76.6  \(\pm\) 1.5\\
       MSTCN-B &  73.5  \(\pm\) 2.4 &  69.3 \(\pm\) 3.0 &  56.6 \(\pm\) 2.4 &  77.0 \(\pm\) 0.9\\
    \hline
  \end{tabular}
  \vspace{-4mm}
\end{table}

\subsection{Action Segmentation}
Table~\ref{tbl:segmetrics} shows the human action segmentation metrics.
We also include a baseline which is not learned, but is derived from the correlation between human and robot actions (Fig.~\ref{fig:activity_corr}), by selecting the most likely human action given the current robot action and progress from the relative occurrences in the training set.
We observe that MSTCN-A is able to better segment human actions, with increased performance for both variants with multimodal inputs.

\begin{table}[b]
  \caption{Robot Generalization (Train$\rightarrow$Test)}
  \label{tbl:hsronly}
  \centering
    \begin{tabular}{l | c | c | c | c}
    \hline
        \textbf{Model} & \textbf{T$\rightarrow$T} & \textbf{T$\rightarrow$K} & \textbf{K$\rightarrow$T} & \textbf{K$\rightarrow$K}\\
        \hline
        I3D-D & 51.1 \(\pm\) 4.1 & 43.0 \(\pm\) 4.4 & 55.1 \(\pm\) 5.5 & 62.2 \(\pm\) 4.6 \\
        MSTCN-A & 70.1 \(\pm\) 2.2 & 56.3 \(\pm\) 3.2 & 47.6 \(\pm\) 6.6 & 48.9 \(\pm\) 3.0 \\
        MSTCN-B & 70.5 \(\pm\) 2.8 & 49.3 \(\pm\) 4.6 & 44.7 \(\pm\) 2.0 & 56.3 \(\pm\) 4.3 \\
    \hline
    \multicolumn{5}{l}{\footnotesize{T: Toyota HSR, K: Kinova Gen3}}
  \end{tabular}
  \vspace{-4mm}
\end{table}

\begin{table}[thpb]
  \caption{Separately trained R2H and H2R networks}
  \label{tbl:h2rvsr2h}
  \centering
    \begin{tabular}{l | c | c }
    \hline
        \textbf{Model} & \textbf{R2H} & \textbf{H2R} \\
        \hline
        I3D-D & 67.2 \(\pm\) 3.3 &  61.5 \(\pm\) 5.8 \\
        MSTCN-A & 69.0 \(\pm\) 5.9 & 65.2 \(\pm\) 6.0 \\
        MSTCN-B & 68.5 \(\pm\) 3.4 & 75.3 \(\pm\) 4.4 \\
    \hline
  \end{tabular}
  \vspace{-4mm}
\end{table}

\subsection{Robot Generalization}
We train the networks using only data from one robot, and evaluate on both to investigate how well the networks are able to generalize to a new robot.
Since F-T and gripper data are represented similarly for both platforms, only the viewpoint, appearance and motions of the robot are the factors which differ between the two.
The results are shown in Table~\ref{tbl:hsronly}.
The performances in all cases are lower than when trained with the full dataset, but we note that the MSTCN variants are able to generalize better when trained on the larger HSR dataset, whereas I3D-D generalizes better when trained on the smaller Kinova dataset.

\subsection{Effect of training on tasks separately}
All previous experiments jointly trained on R2H and H2R.
When the networks are trained on each task separately (see Table~\ref{tbl:h2rvsr2h}) the results are comparable, but slightly lower on average compared to training on both tasks together.

\subsection{Summary}
Overall, the results show that while video is an essential modality for detecting these failures, the addition of F-T in particular, improves the performance.
Adding the robot action segmentation loss for MSTCN-B improves the performance in the unimodal setting, but not in the multimodal setting.
I3D-D, MSTCN-A and MSTCN-B all perform well for the failure detection task, whereas MSTCN-A shows a better performance overall with better segmentation accuracy.
In terms of interpretability of the results, the predicted human actions with the MSTCN networks can be used to analyze the reasons for the failures and when they occur; however this needs to be investigated further in future work.

\section{Limitations}
The handovers in the dataset, including the failures, are performed, and as such may not represent natural handovers or failures; however it is hard to capture naturally occurring failures.
The baseline approaches for failure detection only classify the outcome after the complete trial.
However, a robot should react to a failure as soon as it has occurred, for example, by prompting the person if they have not approached.
Slight modifications of the MSTCN architecture would allow for causal predictions by applying the temporal convolutions only on past inputs, thus enabling online prediction of the outcome.

\section{Conclusions}
In order to address the need for datasets and evaluation methods which consider human-induced failures during handovers, we presented a multimodal handover failure detection dataset containing both R2H and H2R handovers.
The data consists of video, force-torque and robot joint states of both successful and failed handovers. Each video is annotated with the robot's and person's actions during the handover and the overall outcome.
Our baseline approaches show that video is an essential modality for detecting failures and using multimodal data, in particular the force-torque sensor, is beneficial.
Using the human action segmentation approach to detect failures online, and to interpret them based on the human's actions, such that the robot can respond appropriately, are an interesting direction for future research.
%
\bibliographystyle{IEEEtran.bst}
\bibliography{IEEEabrv,references.bib}

\end{document}